%% file: t3dp_arxiv.tex
\documentclass{article} 
\usepackage{arxiv,times}

\input{math_commands.tex}

\usepackage{multirow}
\usepackage{graphicx}
\usepackage[table]{xcolor}

\usepackage[hyperfootnotes=false]{hyperref}
\usepackage{url}
\usepackage{booktabs}
\usepackage{enumitem}
\usepackage{wrapfig}
\usepackage{float}
\usepackage{pifont}
\usepackage{booktabs}
\usepackage{caption}
\usepackage{tabularx}
\usepackage{array}
\usepackage{makecell}
\usepackage{titletoc}
\usepackage{algpseudocode}

\usepackage[ruled,vlined,linesnumbered]{algorithm2e}

\SetKwInput{KwIn}{Input}
\SetKwInput{KwOut}{Output}

\DontPrintSemicolon
\SetAlgoVlined

\SetAlFnt{\small}
\SetAlCapFnt{\small\bfseries}
\SetAlCapNameFnt{\small\bfseries}
\SetAlgoNlRelativeSize{-1}

\SetInd{0.5em}{1.0em}

\algrenewcommand\algorithmicrequire{\textbf{Input:}}
\algrenewcommand\algorithmicensure{\textbf{Output:}}

\newcolumntype{Y}{>{\centering\arraybackslash}X}
\definecolor{avgcolumnblue}{HTML}{E2F3F8}
\definecolor{improvementgreen}{HTML}{008000}
\definecolor{reductionred}{HTML}{C00000}

\title{Text-to-3D Policy: Fine-Grained Language-Behavior Alignment for Unseen Specification Generalization}

\author{
\textbf{Xinhao Yang}$^{1,*}$,
\textbf{Wenhao Wu}$^{1,*}$,
\textbf{Ning Lv}$^{1}$,
\textbf{Yanshen Ding}$^{1}$,
\textbf{Zhenhong Sun}$^{2}$, \\[0.2em]
\textbf{Daoyi Dong}$^{3}$,
\textbf{Chunlin Chen}$^{1}$,
\textbf{Zhi Wang}$^{1}$
\\[0.6em]
$^{1}$Nanjing University
\qquad
$^{2}$Australian National University
\qquad
$^{3}$University of Technology Sydney
\\[0.35em]
\small
\texttt{\{xinhaoyang, wenhaowu, ninglv, yanshending\}@smail.nju.edu.cn}
\\
\small
\texttt{zhenhong.sun@anu.edu.au}
\qquad
\texttt{daoyi.dong@uts.edu.au}
\\
\small
\texttt{\{clchen, zhiwang\}@nju.edu.cn}
}

\newcommand{\res}[2]{#1{\scriptsize$\pm$#2}}

\begin{document}

\maketitle
\begingroup
\renewcommand{\thefootnote}{*}
\footnotetext{Equal contribution.}
\endgroup

\begin{abstract}
3D visuomotor policies provide a strong foundation for spatially precise manipulation, yet current text-to-3D policies struggle to follow unseen fine-grained behavioral specifications beyond those covered by demonstrations. 
We study this challenge as \textit{unseen specification generalization}, where language specifies behaviorally significant variations, such as target position, displacement, or articulated state, that are absent from policy training. 
We find that pretrained language representations and conventional global behavior-language alignment capture coarse task semantics but often blur nearby specifications that require distinct behaviors.
We introduce \textbf{T3DP}, a \textbf{T}ext-to-\textbf{3D} \textbf{P}olicy framework for fine-grained language-behavior alignment. 
Rather than compressing each instruction and demonstration into a single global embedding, T3DP preserves their local structures and establishes bidirectional token-level correspondence between linguistic elements and behavioral segments. 
This directly grounds subtle linguistic variations in the behavior components they affect, preventing closely related specifications from collapsing in the representation space. 
The resulting specification-sensitive language representation conditions a point-cloud-based 3D diffusion policy, enabling more precise control over unseen behavioral specifications without modifying the underlying policy architecture.
Across Meta-World, ManiSkill, and RoboTwin, T3DP improves average held-out-specification success over global language-behavior alignment by \textbf{+11.0-14.2 points}, with gains on all 15 task families; on real-robot tasks, it further raises average success from 47.5\% to 65.0\% (\textbf{+17.5 points}). Representation and action-probe analyses show that fine-grained alignment better preserves specification geometry and action-relevant variation, linking local behavior grounding to downstream control.
\end{abstract}

\section{Introduction}\label{sec:introduction}

\begin{wrapfigure}{r}{0.48\textwidth}
\vspace{-4.1em}
\centering
\includegraphics[width=\linewidth]{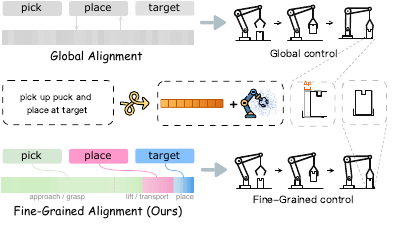}
\caption{\textbf{Global alignment} matches entire sequences, yielding nearly uniform token-to-segment similarities. \textbf{Fine-grained alignment} learns distinct segment-to-token grounding, enabling more precise control with lower errors.
}

\vspace{-2em}
\label{fig:intro}
\end{wrapfigure}

Learning visuomotor policies from 3D observations has emerged as a powerful paradigm for robotic manipulation~\citep{gervet2023act3d}. 
By making scene geometry explicit, point-cloud representations provide strong spatial grounding for precise control and support effective generalization from limited demonstrations~\citep{ze2024dp3}.
Meanwhile, natural language offers a flexible interface for specifying diverse robot behaviors~\citep{wang2026harnesspai}, motivating a new generation of \emph{text-to-3D policies} that translate human instructions into actions grounded in 3D observations~\citep{ke2024diffuseractor}. 
However, a capable text-to-3D policy should determine not only \emph{what} task to perform, but also precisely \emph{how} that task should be executed (Fig.~\ref{fig:intro}).
Whether 3D policies can follow such behaviorally precise instructions beyond the demonstrations observed during training remains an open challenge.

Consider a robot instructed to reach a particular position, push an object by a specified displacement, place it at a desired location, or open an articulated object to a target state.
These instructions describe different \emph{specifications} within the same task family: they share the same coarse task semantics but demand distinct behaviors.
Such specifications often form large, continuous, or compositional spaces, making exhaustive coverage through demonstration collection impractical~\citep{gong2023arnold,zheng2022vlmbench}.
Moreover, the same or similar initial observation may admit multiple valid behaviors, so the intended specification cannot always be inferred from the scene alone and must be communicated through language.
Although prior work has studied generalization to unseen tasks, objects, scenes, and linguistic expressions~\citep{jang2021bcz,shridhar2021cliport}, considerably less attention has been paid to unseen behavior specifications within a task family.
We formalize this challenge as \textbf{unseen specification generalization}: a policy is trained on only a subset of language-specified behaviors and must execute held-out specifications without adaptation.

A natural approach is to condition the policy directly on representations from a pretrained language model~\citep{jang2021bcz}.
These representations encode rich semantic knowledge, but are learned independently of robot interaction and need not preserve how subtle linguistic variations correspond to changes in physical behavior~\citep{zhang2025t2da}.
Recent work therefore grounds language in behavior-relevant representations derived from robot transitions or trajectories~\citep{ma2023liv,myers2023grif,li2024decisionnce}.
Our preliminary study confirms that generic language representations alone provide insufficient actionable specification information. The alignment analysis in Appendix~\ref{app:representation_analysis} further shows that global alignment does not consistently preserve the geometry of continuous specifications.
By compressing an entire instruction and demonstration into a pair of global embeddings, existing objectives primarily emphasize task- or trajectory-level compatibility and provide limited pressure to distinguish nearby specifications.
This raises a critical question: \emph{How can language be grounded in 3D robot behavior without losing the fine-grained structure that distinguishes closely related specifications?}

Our key insight is that fine-grained specifications are often distinguished by \emph{local} rather than global variations.
Closely related instructions share most of their linguistic content, while differing only in behavior-defining elements such as a target, direction, distance, or desired state.
Likewise, their corresponding demonstrations may share most of the motion structure while differing only in the behavioral components affected by the specification.
Global aggregation can obscure these sparse but decisive correspondences.
We therefore hypothesize that directly aligning local linguistic elements
with local behavior representations can better preserve the
specification-dependent structure that global alignment tends to obscure.

Built on this insight, we propose \textbf{T3DP}, a \textbf{T}ext-to-\textbf{3D} \textbf{P}olicy framework for fine-grained language-behavior alignment.
T3DP first encodes demonstration trajectories into a sequence of local behavior representations and preserves the token-level structure of their paired instructions.
Rather than matching only globally pooled features, it establishes bidirectional token-level correspondence between the two modalities, allowing specification-dependent linguistic variations to interact directly with the behavioral components they affect.
This simple alignment objective reshapes the language representation to preserve subtle distinctions among closely related specifications.
The resulting specification-sensitive language representation conditions a point-cloud-based 3D diffusion policy, enabling more precise control over unseen behavioral specifications without redesigning the underlying policy architecture.
Under strictly held-out specification splits, T3DP improves average success over global language-behavior alignment by \textbf{14.2, 13.5, and 11.0 points} on Meta-World, ManiSkill, and RoboTwin, respectively, with gains on all 15 task families. On an AgileX PiperX robot, it raises average success from \textbf{47.5\% to 65.0\%} evaluated over 20 specifications per task. Representation and action-probe analyses further show that its learned condition better preserves specification geometry and action variation.
In summary, our main contributions are:
\begin{itemize}[leftmargin=1.2em, itemsep=0.2em, topsep=0.2em]
    \item We formulate \textbf{unseen specification generalization} for text-to-3D policies, studying whether they can execute fine-grained behavioral specifications absent from policy training.
    \item We identify the coarse granularity of global language-behavior alignment as a key bottleneck and propose \textbf{T3DP}, a simple framework that uses local cross-modal correspondence to learn specification-sensitive language representations for 3D policy learning.
    \item We demonstrate consistent gains across simulation and real-robot tasks, with representation and action-probe analyses showing better preservation of specification structure and action variation.
\end{itemize}

\section{Related Work}
\label{sec:related_work}

\textbf{Language-Conditioned 3D Visuomotor Policies.}
3D visuomotor policies exploit structured geometric representations to provide spatial inductive biases for precise manipulation~\citep{gervet2023act3d,ze2024dp3}.
With advances in large language models~\citep{guo2025deepseek,hu2026diversity,wu2026conflict}, recent work has increasingly explored leveraging their semantic knowledge to guide robot control~\citep{hu2025divide,xie2026look,wu2026trace}.
In 3D visuomotor learning, this trend has led to language-conditioned policies such as PerAct~\citep{shridhar2023peract}, Act3D~\citep{gervet2023act3d}, 3DDA~\citep{ke2024diffuseractor}, and 3D-LOTUS~\citep{garcia2025gembench}, which condition action prediction on linguistic descriptions to enable shared policies across diverse tasks.
Our work considers a finer-grained setting in which instructions share the same task semantics but require distinct behaviors.
We therefore focus on preserving instruction-dependent behavioral variations within a task for downstream control.

\textbf{Generalization to Language-Specified Behaviors.}
Language-conditioned manipulation has studied generalization across unseen tasks, objects, scenes, and linguistic variations~\citep{jang2021bcz,shridhar2021cliport}.
\citet{gong2023arnold} further study unseen continuous goal states within the same task in ARNOLD, primarily varying the desired state along a task-specific dimension.
Our setting additionally considers spatial specifications, such as target positions and relative displacements, where language determines the spatial structure of the required motion.
We study generalization to such held-out specifications within the same task family.

\textbf{Language-Behavior Alignment.}
Multimodal alignment has been widely studied in vision-language learning~\citep{hu2026bridging}, from global image-text contrastive alignment in CLIP~\citep{radford2021clip} to finer-grained correspondence in FILIP~\citep{yao2022filip} and FG-CLIP~\citep{xie2025fgclip}.
Similar ideas have been extended to robot learning, where LIV~\citep{ma2023liv}, GRIF~\citep{myers2023grif}, DecisionNCE~\citep{li2024decisionnce}, and T2DA~\citep{zhang2025t2da} ground language in visual, transition, or behavior representations.
These methods demonstrate the value of language-behavior alignment, but largely operate at the level of a task, transition, or aggregated trajectory.
Fine-grained correspondence benefits local semantic distinctions in vision-language learning, yet remains less explored for grounding specification-bearing language in local robot behavior.
T3DP addresses this gap through local language-behavior correspondence that preserves specification-dependent behavioral variations for downstream control.

\section{Preliminary Study}
\label{sec:preliminary}

\begin{wraptable}{r}{0.42\columnwidth}
\vspace{-2em}
\captionsetup{font=small,skip=2pt}
\caption{Diagnostic study on Meta-World.}
\label{tab:preliminary}
\centering
\footnotesize
\renewcommand{\arraystretch}{0.98}

\begin{tabular*}{\linewidth}{
@{\extracolsep{\fill}}
c c c c
@{}
}
\toprule
\multicolumn{4}{c}{\textbf{(a) Target information}} \\
\midrule
Target & Lang. & Succ. $\uparrow$ & Error $\downarrow$ \\
\midrule
\ding{51} & \ding{51} & 54.7 & 0.093 \\
\ding{51} & \ding{55} & \textbf{55.0} & \textbf{0.091} \\
\ding{55} & \ding{51} & 12.7 & 0.158 \\
\bottomrule
\end{tabular*}

\vspace{0.6em}

\begin{tabular*}{\linewidth}{
@{\extracolsep{\fill}}
l c c c
@{}
}
\multicolumn{4}{c}{\textbf{(b) Specification geometry}} \\
\midrule
Metric & Generic & Global & Ours \\
\midrule
Spearman & 0.373 & 0.591 & \textbf{0.720} \\
kNN@3    & 0.367 & 0.400 & \textbf{0.574} \\
\bottomrule
\end{tabular*}
\vspace{-1em}
\end{wraptable}

\textbf{Language-conditioned control remains strongly tied to visual target cues.}
We conduct a target-cue ablation on Meta-World Reach, where the target can be provided through either observation or language.
As shown in Table~\ref{tab:preliminary}(a), removing the language cue leaves success essentially unchanged (54.7\% vs.\ 55.0\%), whereas removing the visual cue reduces it to 12.7\% and increases the prediction error.
This indicates that language alone provides some actionable information but remains substantially less effective than explicit visual target cues, motivating settings where similar observations require distinct executions.

\textbf{Global alignment only partially preserves fine-grained specification structure.}
We compare generic language representations, global language-behavior alignment, and fine-grained alignment on held-out Reach, Push, and Door Open specifications.
Table~\ref{tab:preliminary}(b) shows that global alignment improves specification geometry over generic representations, while fine-grained alignment further improves both global ordering and local neighborhood preservation.
It suggests that trajectory-level supervision can still obscure distinctions among closely related specifications, highlighting the necessity of our fine-grained alignment.

\section{Method}
\label{sec:method}

Our preliminary study (Section~\ref{sec:preliminary}) shows that generic language representations lack sufficient behaviorally actionable information, while conventional global language-behavior alignment does not consistently preserve the geometry of continuous specifications.
This motivates grounding language in local, specification-relevant behavioral structure rather than relying solely on global task-level summaries.
Building on this insight, we introduce \textbf{T3DP}, which is trained in three stages, as illustrated in Figure~\ref{fig:method}.
Algorithm pseudocodes are presented in Appendix~\ref{app:alg}.
First, we learn a behavior encoder to extract specification-dependent behavior representations from demonstrations (Section~\ref{sec:behavior}).
Second, we ground the text encoder in these representations through fine-grained language-behavior alignment (Section~\ref{sec:alignment}).
Finally, we freeze the aligned text encoder and train a language-conditioned 3D diffusion policy (Section~\ref{sec:policy}).

\begin{figure}[t]
\centering
\includegraphics[width=\textwidth]{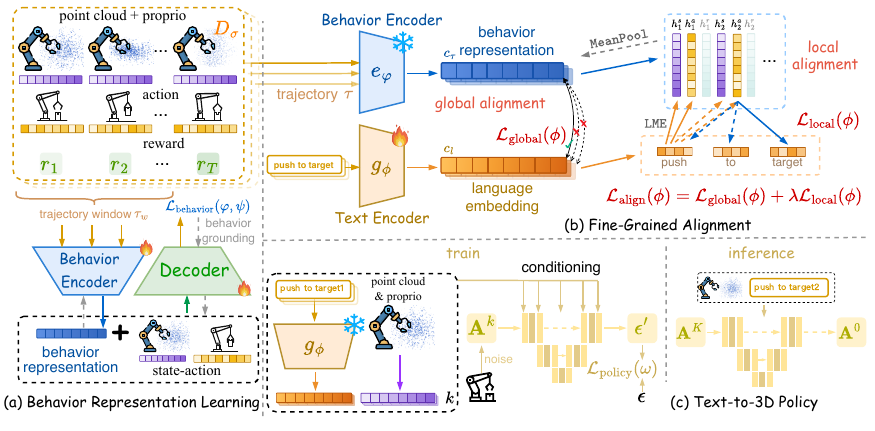}
\caption{\textbf{Overview of T3DP.}
(a) A behavior encoder learns specification-dependent representations from demonstration windows through reward prediction conditioned on state--action pairs.
(b) With the behavior encoder frozen, the text encoder is trained using global language-behavior alignment and local alignment between state--action hidden states and instruction tokens.
(c) The aligned text encoder is frozen to train a language-conditioned 3D diffusion policy using point-cloud observations and proprioceptive states. At inference time, the policy executes unseen specifications from language instructions and current observations, without demonstrations or further adaptation.}
\label{fig:method}
\end{figure}

\subsection{Problem Formulation}
\label{sec:problem}

We consider language-conditioned manipulation with point-cloud observations, where each specification $\sigma$ from a task family $\mathcal{T}$ defines an MDP $\mathcal{M}_{\sigma}=(\mathcal{S},\mathcal{A},\mathcal{P},r_{\sigma},\gamma)$.
Specifications in a family share $\mathcal{S}$, $\mathcal{A}$, and $\mathcal{P}$, but differ in the specification-dependent reward $r_{\sigma}$; each state contains a point-cloud observation $o$ and proprioceptive state $p$.
Training uses paired instructions $l_{\sigma_n}$ and expert datasets $\mathcal{D}_{\sigma_n}=\{\tau_{\sigma_n}^{(k)}\}_{k=1}^{K_n}$ for specifications $\{\sigma_n\}_{n=1}^{N}$, with $\tau=\{(s_t,a_t,r_t)\}_{t=1}^{T}$, where $N$, $K_n$, and $T$ denote the number of training specifications, expert trajectories for specification $\sigma_n$, and the trajectory length, respectively.
At test time, the policy receives only the language instruction for a held-out specification from the same family.
Our goal is to learn a behavior-grounded language encoder that distinguishes related instructions and a policy $\pi(a_t\mid o_t,l)$ that generalizes zero-shot.

\subsection{Behavior Representation Learning from Demonstrations}
\label{sec:behavior}

Table~\ref{tab:preliminary} shows that generic language representations alone do not carry enough behaviorally actionable information to specify a task.
We therefore first learn a behavior encoder that extracts specification-dependent behavioral structure from demonstrations before aligning it with language.

Specifications within the same task family share the same state and action spaces and transition dynamics, and differ only in the specification-dependent reward $r_{\sigma}$.
The behavioral differences that separate closely related specifications therefore originate from the reward rather than from dynamics or embodiment.
This motivates training the behavior encoder with a reward-predictive objective: the behavior embedding of a trajectory segment should be able to predict the reward corresponding to any state-action pair within the same specification.
Succeeding at this requires the embedding to capture specification-dependent behavioral structure, rather than surface-level motion statistics.

Given a length-$L$ window $\tau_w = \{(s_t,a_t,r_t)\}_{t=w}^{w+L-1}$, $e_{\varphi}$ uses PointNet and modality-specific MLPs to tokenize each timestep, followed by a bidirectional Transformer and mean pooling.
The point cloud and proprioceptive state form $\tilde{s}_t$, while the action and reward form $\tilde{a}_t$ and $\tilde{r}_t$.
All tokens at same timestep share the same temporal positional encoding.
The resulting $3L$ tokens are arranged in temporal order and compute the behavior embedding as
\begin{align}
\mathbf{X}_w
&=
\big[
\tilde{s}_w,\tilde{a}_w,\tilde{r}_w,\ldots,
\tilde{s}_{w+L-1},\tilde{a}_{w+L-1},\tilde{r}_{w+L-1}
\big], \\
e_{\varphi}(\tau_w)
&=
\operatorname{MeanPool}\!\left(
\operatorname{Transformer}(\mathbf{X}_w)
\right).
\end{align}

To train $e_{\varphi}$, a lightweight decoder $d_{\psi}$ predicts the reward $\hat{r}=d_{\psi}\!\left(e_{\varphi}(\tau_w),s,a\right)$ for a state-action pair drawn from the same specification as the encoded trajectory.
We select a trajectory $\tau\in\mathcal{D}_{\sigma}$, a segment $\tau_w\subset\tau$, and a transition $(s,a,r)\in\tau\setminus\tau_w$, ensuring that the target reward is not contained in the encoded segment, and jointly optimize the encoder and decoder with
\begin{equation}
\mathcal{L}_{\mathrm{behavior}}(\varphi,\psi)
=
\mathbb{E}_{\substack{
\sigma\sim\mathcal{T},\;\tau\in\mathcal{D}_{\sigma},\;\tau_w\subset\tau\\
(s,a,r)\in\tau\setminus\tau_w
}}
\!\left[
\left(
d_{\psi}\!\left(e_{\varphi}(\tau_w),s,a\right)-r
\right)^2
\right].
\end{equation}

\subsection{Fine-Grained language-behavior alignment}
\label{sec:alignment}

The behavior encoder captures what each specification entails in physical execution, but demonstrations are generally unavailable for novel specifications at deployment.
Language provides a scalable interface, but generic representations do not necessarily encode behavioral consequences.
Language-behavior alignment transfers actionable structure from demonstrations into instructions, which is crucial when a single observation allows for multiple plausible behaviors and generic language lacks distinction.
It is also important for multimodal policy learning because precise point-cloud observations can otherwise dominate generic language features, as shown in our preliminary study.
We therefore seek an alignment that preserves distinctions among closely related specifications and provides an effective condition for the policy below.

\paragraph{Limitations of global alignment.}
A common approach to behavior grounding is to align globally pooled instruction and behavior embeddings~\citep{radford2021clip,zhang2025t2da}.
Although global alignment improves over generic language representations (Appendix~\ref{app:representation_analysis}), its gains are limited and inconsistent for closely related specifications.
Such specifications share most linguistic and behavioral content, differing only in decisive elements such as target location, direction, or desired state.
Pooling both modalities can obscure these differences, motivating finer-grained alignment.

\paragraph{Fine-grained alignment.}
To preserve these subtle distinctions, we retain local language and behavior representations and explicitly model their cross-modal correspondence.
This grounds specification-defining linguistic elements in the behavioral components they affect and complements global alignment by preserving behaviorally consequential changes.

\paragraph{Implementation and loss function.}
We use a pretrained Transformer text encoder $g_{\phi}$, similar to T5 and CLIP~\citep{raffel2020t5,radford2021clip}.
For an instruction $l$, the text encoder produces token representations $E(l)=g_{\phi}(l)=\{e_j\}_{j=1}^{M}$ and a pooled representation $c_l=\frac{1}{M}\sum_{j=1}^{M}e_j$.
For its paired full trajectory $\tau=\{(s_t,a_t,r_t)\}_{t=1}^{T}$, let $\mathbf{H}_{\varphi}(\tau)=\big[h_1^s,h_1^a,h_1^r,\ldots,h_T^s,h_T^a,h_T^r\big]$ denote the output of the trained behavior Transformer before mean pooling.
We use the pooled representation $c_\tau=e_\varphi(\tau)=\operatorname{MeanPool}(\mathbf{H}_\varphi(\tau))$ for global alignment and retain the state and action hidden states $Z(\tau)=\{z_i\}_{i=1}^{2T}=\big[h_1^s,h_1^a,\ldots,h_T^s,h_T^a\big]$ for fine-grained alignment.

Given a batch of $B$ paired trajectories and instructions, global alignment uses $c_\tau$ and $c_l$ in a contrastive objective with matched pairs as positives and mismatched pairs as negatives:
\begin{equation}
\mathcal{L}_{\mathrm{global}}(\phi)
=
-\frac{1}{2B}\sum_{b=1}^{B}
\left[
\log
\frac{\exp(c_{\tau_b}^{\top}c_{l_b}/\gamma)}
{\sum_{q=1}^{B}\exp(c_{\tau_b}^{\top}c_{l_q}/\gamma)}
+
\log
\frac{\exp(c_{\tau_b}^{\top}c_{l_b}/\gamma)}
{\sum_{q=1}^{B}\exp(c_{\tau_q}^{\top}c_{l_b}/\gamma)}
\right],
\end{equation}
where $\gamma$ is the contrastive temperature.
This brings paired embeddings closer and separates unpaired ones within the batch.

Because global pooling may dilute specification-defining elements, we additionally match state-action hidden states with instruction tokens.
For each segment-token pair, we compute its cosine similarity $s_{ij}=\operatorname{cos}(z_i,e_j)$ and define two directional late-interaction scores:
\begin{equation}
S_{\tau\rightarrow l}(Z,E)
=
\frac{1}{N}\sum_{i=1}^{N}
\operatorname{LME}_{\beta}\big(\{s_{ij}\}_{j=1}^{M}\big),
\qquad
S_{l\rightarrow\tau}(Z,E)
=
\frac{1}{M}\sum_{j=1}^{M}
\operatorname{LME}_{\beta}\big(\{s_{ij}\}_{i=1}^{N}\big),
\end{equation}

where $\operatorname{LME}_{\beta}(\{x_k\}_{k=1}^{K})=\frac{1}{\beta}\log\left(\frac{1}{K}\sum_{k=1}^{K}\exp(\beta x_k)\right)$ is a smooth approximation to the maximum, with $\beta>0$ controlling the matching sharpness.
The two scores softly match segments to relevant tokens and tokens to the segments expressing their behavioral meaning.
We apply the same bidirectional contrastive construction to these fine-grained scores:
\begin{equation}
\mathcal{L}_{\mathrm{local}}(\phi)
=
-\frac{1}{2B}\sum_{b=1}^{B}
\left[
\log
\frac{\exp(S_{\tau\rightarrow l}(Z_b,E_b)/\gamma)}
{\sum_{q=1}^{B}\exp(S_{\tau\rightarrow l}(Z_b,E_q)/\gamma)}
+
\log
\frac{\exp(S_{l\rightarrow\tau}(Z_b,E_b)/\gamma)}
{\sum_{q=1}^{B}\exp(S_{l\rightarrow\tau}(Z_q,E_b)/\gamma)}
\right].
\end{equation}
The final objective combines global and local alignment with weight $\lambda$:
\begin{equation}
\mathcal{L}_{\mathrm{align}}(\phi)
=
\mathcal{L}_{\mathrm{global}}(\phi)
+\lambda\mathcal{L}_{\mathrm{local}}(\phi).
\end{equation}
During alignment, we freeze the behavior encoder $e_{\varphi}$ and fine-tune the text encoder using LoRA.

\subsection{Language-Guided 3D Policy Learning}
\label{sec:policy}

Natural language provides a universal interface compatible with foundation models and expresses semantic or compositional goals that fixed numerical parameters cannot fully capture, such as ``open the door''.
We instantiate the downstream policy with a point-cloud-based 3D diffusion policy~\citep{ze2024dp3}, which encodes the current point-cloud observation $o_t$ and proprioceptive state $p_t$ into a compact 3D feature and combines them with language to denoise an action sequence.
Because the observation specifies scene geometry but not the intended behavior when one scene admits multiple specifications, the policy requires complementary, specification-dependent language guidance.

Given an instruction $l$, we use the behavior-aligned text encoder from Section~\ref{sec:alignment} to compute pooled language embedding $c_l$ as the policy condition.
Although the downstream policy consumes only this pooled representation, language-behavior alignment reshapes $c_l$ to preserve fine-grained behavioral distinctions.
We freeze $g_{\phi}$ during policy learning and train a single policy for all training specifications within each task family $\mathcal{T}$.
Let $\mathbf{A}^{0}$ denote an expert action sequence and $\mathbf{A}^{k}$ its noisy version at diffusion step $k$.
The denoising network $\epsilon_{\omega}$ is conditioned on the point-cloud observation, proprioceptive state, and aligned instruction representation, and is optimized using
\begin{equation}
\mathcal{L}_{\mathrm{policy}}(\omega)
=
\mathbb{E}_{\mathbf{A}^{0},\,\epsilon,\,k}
\left[
\left\|
\epsilon-
\epsilon_{\omega}
\left(
\mathbf{A}^{k},k,o_t,p_t,c_l
\right)
\right\|_{2}^{2}
\right],
\end{equation}
where $\epsilon\sim\mathcal{N}(0,I)$ denotes the noise added in the forward diffusion process.
Importantly, specification-dependent targets, such as desired positions, are not rendered in the point-cloud observation because such information is generally unavailable in real scenes and must instead be communicated through language.
At inference time, the policy requires only the current observation, robot state, and instruction to execute held-out specifications without demonstrations or further adaptation.

\section{Experiments}
\label{sec:experiments}

Our experiments are organized around four questions.
First, does T3DP consistently improve generalization to unseen task
specifications over competing baselines
(Section~\ref{sec:main_experiments})
Second, we isolate the contribution of fine-grained alignment by comparing
it with raw language conditioning and global language-behavior alignment
(Section~\ref{sec:ablation}).
Third, we examine whether the advantage persists with denser specification
coverage and joint training across heterogeneous task families
(Section~\ref{sec:scalability}).
Finally, we evaluate whether fine-grained alignment transfers to a physical
robot (Section~\ref{sec:real_world}).
Additional RoboTwin results and representation analyses are provided in
Appendix~\ref{app:additional_results}.

\textbf{Benchmarks.}
We evaluate T3DP on three robot manipulation benchmarks: Meta-World~\citep{yu2021metaworld}, ManiSkill~\citep{mu2021maniskill}, and RoboTwin~\citep{chen2026robotwin}.
Together, these benchmarks cover diverse task families, including spatial goal reaching, object interaction, placement, and articulated-object manipulation.
For each task family, we construct disjoint training and test specification sets, and evaluate each baseline on the test specification set without further adaptation.
The desired behavior is explicitly specified through language, yet such specifications may be semantically similar to one another and therefore hard to distinguish from raw instruction embeddings alone.
Detailed benchmark and task descriptions are provided in Appendix~\ref{app:task_details}.

\textbf{Baselines.}
We compare T3DP with baselines from three categories:
i) 3D diffusion policy~\citep{ze2024dp3}, a strong 3D visuomotor policy without language conditioning;
ii) independent language-conditioned 3D policies, including 3DDA~\citep{ke2024diffuseractor} and 3D-LOTUS~\citep{garcia2025gembench};
and iii) behavior--instruction alignment methods, represented by T2DA~\citep{zhang2025t2da}, which performs global language-behavior alignment.
In contrast, T3DP introduces fine-grained local alignment between behavior and language representations.
Following the data regime of DP3, all methods are trained with only a limited number of expert demonstrations.
All methods use the same demonstrations, observation inputs, task instructions, training and test specification splits, and evaluation protocol.
Detailed baseline implementations are provided in Appendix~\ref{app:baselines}.

\subsection{Main Results}
\label{sec:main_experiments}

\begin{table}[t]
\centering
\caption{Unseen-specification generalization on Meta-World and ManiSkill. We report the average success rate (\%) and corresponding standard deviation over three independently trained seeds. ``Avg.'' denotes the mean across the five task families within each benchmark. Best and second-best results are shown in \textbf{bold} and \underline{underlined}, respectively.}
\label{tab:main_results}

\small
\renewcommand{\arraystretch}{1.03}
\setlength{\tabcolsep}{3pt}

\begin{tabular}{l*{5}{c}>{\columncolor{avgcolumnblue}}c}
\toprule
\multicolumn{7}{c}{\textbf{Meta-World}} \\
\midrule
\textbf{Method}
& \textbf{Reach}
& \textbf{Push}
& \textbf{Pick Place}
& \textbf{Door Open}
& \textbf{Drawer Open}
& \textbf{Avg.} \\
\midrule

DP3~\citep{ze2024dp3} & \res{9.7}{2.3} & \res{19.3}{4.0} & \res{12.7}{9.3} & \res{15.3}{5.0} & \res{24.0}{4.4} & 16.2 \\
3DDA~\citep{ke2024diffuseractor} & \res{23.3}{18.9} & \res{31.7}{20.2} & \res{26.7}{11.5} & \res{16.7}{2.9} & \res{6.7}{7.6} & 21.0 \\
3D-LOTUS~\citep{garcia2025gembench} & \res{21.7}{7.6} & \res{1.7}{2.9} & \res{1.7}{2.9} & \res{38.3}{15.3} & \res{\underline{28.3}}{2.9} & 18.3 \\
T2DA~\citep{zhang2025t2da} & \res{\underline{24.3}}{4.0} & \res{\underline{45.7}}{8.6} & \res{\underline{46.0}}{1.7} & \res{\underline{41.3}}{8.3} & \res{23.3}{2.1} & \underline{36.1} \\
\textbf{T3DP (Ours)} & \res{\textbf{38.3}}{2.9} & \res{\textbf{64.7}}{17.9} & \res{\textbf{58.0}}{2.6} & \res{\textbf{59.7}}{4.2} & \res{\textbf{30.7}}{12.7} & \textbf{50.3} \\
\midrule
$\Delta$ vs. DP3 & \textcolor{improvementgreen}{+28.6} & \textcolor{improvementgreen}{+45.4} & \textcolor{improvementgreen}{+45.3} & \textcolor{improvementgreen}{+44.4} & \textcolor{improvementgreen}{+6.7} & \textcolor{improvementgreen}{+34.1} \\

\midrule
\multicolumn{7}{c}{\textbf{ManiSkill}} \\
\midrule
\textbf{Method}
& \textbf{Reach}
& \textbf{Push Cube}
& \textbf{Pull Cube}
& \textbf{Pick Cube}
& \textbf{Turn Valve}
& \textbf{Avg.} \\
\midrule

DP3~\citep{ze2024dp3} & \res{3.3}{0.6} & \res{8.0}{2.0} & \res{16.3}{4.2} & \res{53.3}{1.5} & \res{21.0}{3.0} & 20.4 \\
3DDA~\citep{ke2024diffuseractor} & \res{35.0}{8.7} & \res{36.7}{25.2} & \res{6.7}{2.9} & \res{8.3}{5.8} & \res{\textbf{56.7}}{7.6} & 28.7 \\
3D-LOTUS~\citep{garcia2025gembench} & \res{\textbf{41.7}}{2.9} & \res{\textbf{51.7}}{5.8} & \res{35.0}{8.7} & \res{0.0}{0.0} & \res{\underline{53.3}}{2.9} & 36.3 \\
T2DA~\citep{zhang2025t2da} & \res{32.3}{4.6} & \res{15.0}{9.1} & \res{\underline{48.3}}{6.1} & \res{\underline{72.0}}{12.5} & \res{36.0}{2.0} & \underline{40.7} \\
\textbf{T3DP (Ours)} & \res{\underline{38.0}}{4.4} & \res{\underline{47.3}}{4.0} & \res{\textbf{58.7}}{3.5} & \res{\textbf{77.7}}{6.0} & \res{49.3}{1.2} & \textbf{54.2} \\
\midrule
$\Delta$ vs. DP3 & \textcolor{improvementgreen}{+34.7} & \textcolor{improvementgreen}{+39.3} & \textcolor{improvementgreen}{+42.4} & \textcolor{improvementgreen}{+24.4} & \textcolor{improvementgreen}{+28.3} & \textcolor{improvementgreen}{+33.8} \\

\bottomrule
\end{tabular}
\vspace{-1em}
\end{table}

We evaluate unseen-specification generalization on Meta-World and ManiSkill.
Policies are trained independently for each task family with different specifications.
Table~\ref{tab:main_results} reports the results of comparing T3DP against several baselines, with all results averaged over three training seeds.

\textbf{Meta-World.}
T3DP achieves the highest average success at \textbf{50.3\%}, improving on T2DA, the strongest baseline, by \textbf{14.2 percentage points}.
The improvement is consistent across all five task families, with gains ranging from 7.4 to 19.0 points.
On the more challenging Push and Pick Place tasks, which require precise object interaction and placement control, several baselines achieve substantially lower success rates.
T3DP delivers particularly pronounced gains on these tasks, suggesting that fine-grained alignment enables instruction representations to capture the subtle distinctions between closely related behavioral specifications.

\textbf{ManiSkill.}
Across the five task families, T3DP achieves an average success rate of \textbf{54.2\%}, outperforming T2DA by \textbf{13.5 points}, with gains on every task.
The largest improvement appears on Push Cube, where T3DP reaches 47.3\% success compared with 15.0\% for T2DA.
On Pick Cube, T3DP achieves 77.7\%, compared with 8.3\% for 3DDA and 0.0\% for 3D-LOTUS.
These results show the benefit of fine-grained alignment for distinguishing closely related behavioral specifications.

Taken together, the Meta-World and ManiSkill results demonstrate that T3DP maintains its advantage across distinct benchmarks and manipulation settings under limited supervision.
We further assess its transferability on RoboTwin, where T3DP improves over global alignment across five additional task families; detailed results are provided in Appendix~\ref{app:robotwin_results}.

\subsection{Ablation Experiments}
\label{sec:ablation}

\begin{wrapfigure}{r}{0.48\columnwidth}
\vspace{-4em}
\centering
\includegraphics[width=\linewidth]{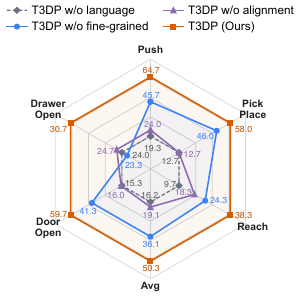}
\caption{
Hierarchical component ablation on Meta-World.
Each spoke is scaled independently with the task-specific T3DP w/o language
score as its inner reference and additional range above the maximum observed
score. Avg denotes the mean over the five task families.
}
\label{fig:ablation}
\vspace{0.5em}
\centering
\includegraphics[width=\linewidth]{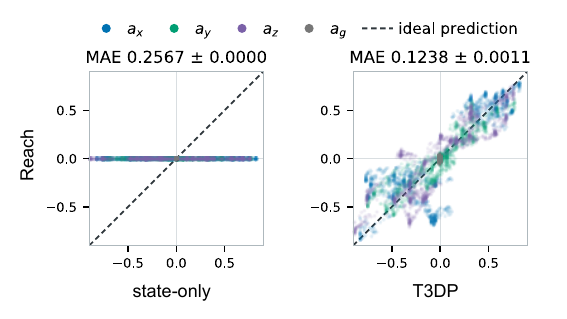}
\caption{
With the robot state fixed and only the language specification varied, the state-only predictor collapses to the average action, whereas T3DP's fine-grained representation follows each specification's target action.
}
\label{fig:action_probe}
\vspace{-3.0em}
\end{wrapfigure}

We isolate the contribution of T3DP's language-conditioning and alignment components on all five Meta-World task families.
The four variants form a hierarchical comparison: \texttt{T3DP w/o language} removes the language input and is equivalent to DP3; \texttt{T3DP w/o alignment} directly uses the pretrained language representation without language-behavior alignment; \texttt{T3DP w/o fine-grained} retains only the global pooled alignment used by T2DA; and \texttt{T3DP} adds the proposed token-to-segment fine-grained alignment.
Results are shown as success rates for each task family and their average in Figure~\ref{fig:ablation}.

\textbf{Does T3DP's alignment contribute to the improvement?}
Adding raw language conditioning to the no-language baseline increases average success from 16.2\% to 19.1\%, a gain of 2.9 points.
Global language-behavior alignment raises the average to 36.1\%, demonstrating the benefit of behavior grounding.
Adding local fine-grained alignment (T3DP) achieves the strongest average success at \textbf{50.3\%}, improving over the global-only variant by \textbf{14.2 points}.
T3DP also performs best on every task family, with gains of 7.4--19.0 points over the global-only variant.
Together, these results show that fine-grained language-behavior alignment consistently benefits over raw language conditioning and global alignment.

\textbf{Does the learned language representation encode behavior-relevant information?}
We train a small predictor to recover the expert action from the learned language condition, holding the robot state fixed and varying only the language specification.
A predictor given only the state cannot distinguish actions from different specifications and collapses to their average action, whereas one given T3DP's fine-grained language representation follows each specification's target action (Figure~\ref{fig:action_probe}).
On Reach, T3DP lowers the mean absolute action error (MAE) from 0.2567 to 0.1238, beating global alignment (T2DA, 0.1423).
The full results on Reach and Push are in Appendix~\ref{app:representation_analysis}.

\vspace{-0.3em}
\subsection{Scalability Analysis}
\label{sec:scalability}
\vspace{-0.3em}

We further examine whether the advantage of fine-grained language-behavior alignment persists as the learning problem scales along two complementary dimensions: denser coverage of the specification space and joint training across heterogeneous task families.

\textbf{Scaling with specification coverage.}
We vary the number of training specifications per Meta-World task among 10, 20, and 40, while keeping the held-out test set and evaluation protocol fixed.
Each specification contributes one expert demonstration, so increasing the scale expands specification coverage without changing demonstrations per specification.

\begin{table}[t]
    \centering
    \caption{
    Effect of training specification coverage on unseen-specification generalization in Meta-World.
    Each training specification contains one expert demonstration, and success rates (\%) are evaluated on a common held-out specification set.
    }
    \label{tab:specification_scale_ablation}
    \small
    \setlength{\tabcolsep}{6pt}
    \begin{tabular}{lcccccc>{\columncolor{avgcolumnblue}}c}
        \toprule
        Method
        & \# Train Specs.
        & Reach
        & Push
        & Pick Place
        & Door Open
        & Drawer Open
        & Avg. \\
        \midrule
        T2DA
        & 10 & 22.3 & 25.3 & 26.7 & 28.7 & 16.7 & 23.9 \\
        & 20 & 24.3 & 45.7 & 46.0 & 41.3 & 23.3 & 36.1 \\
        & 40 & 36.1 & 63.3 & 75.0 & 100.0 & 71.7 & 69.2 \\
        \midrule
        \textbf{T3DP (Ours)}
        & 10 & 30.3 & 37.0 & 34.0 & 35.3 & 25.3 & 32.4 \\
        & 20 & 38.3 & 64.7 & 58.0 & 59.7 & 30.7 & 50.3 \\
        & 40 & 47.2 & 75.0 & 76.7 & 100.0 & 83.3 & 76.4 \\
        \bottomrule
    \end{tabular}
\vspace{-0.5em}
\end{table}

Across all three specification scales, T3DP consistently outperforms global alignment method T2DA.
With 10, 20, and 40 training specifications, T3DP raises the average success rate from 23.9\%, 36.1\%, and 69.2\% to 32.4\%, 50.3\%, and 76.4\%, respectively.
Both methods improve as the training specifications cover more of the specification space, while T3DP maintains an advantage at every scale.
This consistent advantage across coverage levels indicates that fine-grained alignment remains effective as the number of training specifications increases.

\textbf{Joint training across task families.}
To evaluate whether fine-grained alignment also transfers to a more heterogeneous learning setting, we jointly train a single policy on all five Meta-World task families.
These families span spatial targets, relative displacements, object placement, and articulated states.
The demonstrations and train--test specification splits remain unchanged, with demonstrations from all task families combined into one shared training set.

\begin{table}[t]
    \centering
    \caption{
    Joint multi-task training on Meta-World.
    A single policy is trained jointly across all five task families and
    evaluated on held-out specifications within each family.
    Success rates (\%) are averaged over three seeds.
    Best results are shown in \textbf{bold}.
    }
    \label{tab:metaworld_mixed_training}
    \small
    \setlength{\tabcolsep}{6pt}
    \begin{tabular}{lccccc>{\columncolor{avgcolumnblue}}c}
        \toprule
        Method
        & Push
        & Pick Place
        & Reach
        & Door Open
        & Drawer Open
        & Avg. \\
        \midrule
        DP3
        & 14.3
        & 12.0
        & 1.0
        & 10.3
        & 18.7
        & 11.3 \\
        T2DA
        & 42.7
        & \textbf{53.0}
        & \textbf{36.3}
        & 31.0
        & 14.3
        & 35.5 \\
        \textbf{T3DP (Ours)}
        & \textbf{56.7}
        & 42.7
        & 35.3
        & \textbf{47.3}
        & \textbf{21.3}
        & \textbf{40.7} \\
        \bottomrule
    \end{tabular}
\vspace{-1em}
\end{table}

Under joint training across diverse task families, T3DP improves average success by 5.2 points over T2DA and 29.4 points over DP3.
T3DP performs best on Push, Door Open, and Drawer Open, while T2DA is better on Pick Place and slightly better on Reach.
Overall, T3DP remains effective in the mixed-task setting, showing that fine-grained alignment transfers to a single policy across heterogeneous task families.

\vspace{-0.3em}
\subsection{Real-world Transfer}
\label{sec:real_world}
\vspace{-0.3em}

\begin{figure}[t]
\centering
\includegraphics[width=0.98\columnwidth]{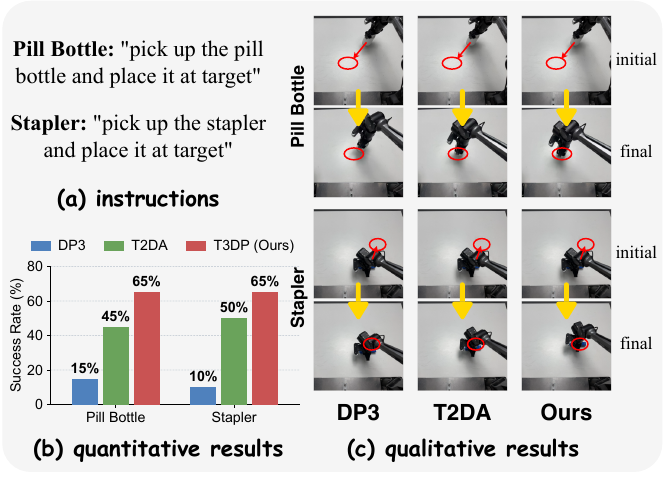}
\caption{Real-world manipulation experiments on a PiPer X robotic arm. (a) Language instructions for two task families \textbf{Pill Bottle} and \textbf{Stapler}. (b) Quantitative success rates for DP3, T2DA, and T3DP. (c) Qualitative real-world control results showing the final object positions relative to the marked targets. T3DP achieves a higher success rate and higher final placement precision.}
\label{fig:real_world}
\vspace{-1em}
\end{figure}

We further evaluate whether the learned language-behavior correspondence transfers to a physical robot.
We use a PiPer X robotic arm and construct the point cloud exclusively from a fixed D435i depth camera mounted above the workspace.
We construct two task families, both require moving an object from its initial location to a designated target location, as illustrated in Figure~\ref{fig:real_world}.
A trial is successful when the final object position is within 3\,cm of the target.
Each task family is evaluated on 20 distinct specifications.
Additional details of the real-world setup are provided in Appendix~\ref{app:real_robot}.

We compare DP3 without language conditioning, T2DA with global language-behavior alignment, and T3DP with the proposed fine-grained alignment under the same task and evaluation settings.

\textbf{Quantitatively},
global language–behavior alignment substantially improves the success rate of DP3 without language conditioning across both tasks. Adding fine-grained alignment on top of it yields a further clear gain, with T3DP achieving the highest success rates across both tasks.
These results show that behavior grounding provides useful language guidance in the real-world setting, while fine-grained correspondence offers additional benefits beyond global alignment.
\textbf{Qualitatively},
DP3 produces final object positions farthest from the target locations.
Although T2DA brings the objects close to their targets, its placement remains less precise than that of T3DP.

Together, the quantitative and qualitative results indicate that fine-grained alignment improves both the reliability and the precision of language-conditioned manipulation beyond simulation.

\vspace{-0.4em}
\section{Conclusion, Limitations and Future Work}
\vspace{-0.4em}
\label{sec:conclusion}

We study unseen specification generalization in text-to-3D policies and introduce T3DP, which aligns instruction tokens with local behavior representations learned from demonstrations.
This fine-grained alignment produces specification-sensitive conditions for a 3D diffusion policy without demonstrations at inference time.
T3DP improves held-out specification execution across simulation benchmarks and real-robot tasks, and representation analyses show better preservation of specification structure and action variation.
T3DP requires expert trajectories with reward signals, however, and our evaluation is limited to held-out specifications within known task families under templated instructions.
Further limitations are discussed in Appendix~\ref{app:limitations}.
Future work could learn specification-dependent behavior representations directly from state-action sequences and extend T3DP to unseen task families, free-form instructions, and diverse embodiments.

\clearpage

\bibliographystyle{iclr2027_conference}
\bibliography{references}

\clearpage
\appendix

\part*{Appendix}
\addcontentsline{toc}{part}{Appendix}
\startcontents[appendix]
\printcontents[appendix]{}{1}{}
\clearpage

\section{Task, Language Specifications, and Dataset Splits}
\label{app:task_details}
\label{app:setup}

\begin{list}{\textbullet}{%
  \settowidth{\labelwidth}{\textbullet}%
  \setlength{\labelsep}{0.5em}%
  \setlength{\leftmargin}{\labelwidth}%
  \addtolength{\leftmargin}{\labelsep}%
}

\item \textbf{Meta-World.}
We evaluate \emph{Push}, \emph{Pick Place}, \emph{Reach}, \emph{Door Open},
and \emph{Drawer Open}. Each observation consists of 512 XYZ points and a
9-dimensional proprioceptive state, and the policy predicts a 4-dimensional
continuous action. The numerical task specification is excluded from the
observation and provided only through language. Reach parameterizes the
gripper target location; Push parameterizes a relative planar displacement;
Pick Place parameterizes the object target location; and Door Open and
Drawer Open parameterize normalized articulation progress over a restricted
feasible range. Each rollout is capped at 200 environment steps.

\item \textbf{RoboTwin.}
We evaluate \emph{Open Microwave}, \emph{Open Cabinet Drawer},
\emph{Move Pill Bottle}, \emph{Move Playing Card}, and \emph{Move Stapler}.
The policy receives 512 XYZ points and a 14-dimensional robot joint state and
predicts 14-dimensional joint-position actions. The articulated tasks
parameterize target opening progress, while the remaining tasks parameterize
planar placement locations sampled over their feasible workspaces. Episode
horizons are 550 steps for \emph{Open Microwave}, 410 for
\emph{Open Cabinet Drawer}, and 200 for the placement tasks. To control
initial-state variation, each task uses 20 reproducibly sampled initial object
configurations around its nominal pose.

\item \textbf{ManiSkill.}
We evaluate \emph{Reach}, \emph{Push Cube}, \emph{Pull Cube},
\emph{Pick Cube}, and \emph{Turn Valve}. Each observation contains 512 XYZ
points and a 32-dimensional proprioceptive state, while the numerical task
specification is provided only through language. The policy predicts a
7-dimensional end-effector delta-pose action. The five domains parameterize,
respectively, a Cartesian target, a forward displacement, a planar pulling
target, a lifting extent, and a rotation extent. In \emph{Pull Cube}, the
expert grasps the cube, moves it to the specified planar target, and releases
it.

\end{list}

Unless otherwise specified, simulation benchmark success rates are averaged
over three independently trained seeds. Real-robot results use 20 held-out
trials per task and method, while action-probe results are averaged over four
trajectory splits.

\subsection{Dataset Splits}
\label{app:dataset_splits}

All 15 task domains appear in both training and evaluation. Generalization is
therefore measured over \emph{held-out task specifications}, rather than
unseen task categories. For each domain, we use 20 training specifications
and 20 held-out test specifications, with one successful demonstration
associated with each training specification.

Training and test specifications are disjoint within each task and are sampled
to provide approximately uniform coverage of the corresponding feasible
specification space. Test specifications are not used during representation
alignment or policy training. Any optimization-time validation split is
constructed only from the training demonstrations and is excluded from the
reported held-out evaluation.

\subsection{Language Instructions and Task Specifications}
\label{app:language_specs}

We represent each task instance by a specification
\begin{equation}
    \sigma = (f_1, f_2, \ldots, f_d),
\end{equation}
where each factor corresponds to a behaviorally relevant property of the
desired outcome, such as a target location, displacement, placement goal, or
articulation extent.
The specification is expressed through a task-specific natural language
template whose values are instantiated from $\sigma$.

For each task, training and test instructions follow the same linguistic
construction, while their underlying specifications are disjoint.
Our evaluation therefore focuses on \emph{generalization to unseen behavioral
specifications}, rather than generalization to paraphrased language.
Table~\ref{tab:app-task-language} summarizes the task specifications and
instruction templates used in each benchmark.

\begin{table}[t]
\centering
\caption{
Task specifications and language instruction templates across the evaluated benchmarks.
}
\label{tab:app-task-language}
\small
\setlength{\tabcolsep}{5pt}
\renewcommand{\arraystretch}{1.10}

\begin{tabularx}{\linewidth}{
    @{}
    >{\raggedright\arraybackslash}p{0.22\linewidth}
    >{\raggedright\arraybackslash}p{0.30\linewidth}
    >{\raggedright\arraybackslash}X
    @{}
}
\toprule
\multicolumn{3}{c}{\textbf{Meta-World}} \\
\midrule
\textbf{Task}
& \textbf{Specification}
& \textbf{Instruction Template} \\
\midrule

Push
&  Relative displacement
& ``Push the object [lateral displacement] and [forward displacement].'' \\

Pick Place
& Placement target location
& ``Pick up the object and place it at [target location].'' \\

Reach
& End-effector target location
& ``Move the gripper to [target location].'' \\

Door Open
& Target opening extent
& ``Open the door to [opening extent].'' \\

Drawer Open
& Target opening extent
& ``Open the drawer to [opening extent].'' \\

\bottomrule
\end{tabularx}

\vspace{0.7em}

\begin{tabularx}{\linewidth}{
    @{}
    >{\raggedright\arraybackslash}p{0.22\linewidth}
    >{\raggedright\arraybackslash}p{0.30\linewidth}
    >{\raggedright\arraybackslash}X
    @{}
}
\toprule
\multicolumn{3}{c}{\textbf{RoboTwin}} \\
\midrule
\textbf{Task}
& \textbf{Specification}
& \textbf{Instruction Template} \\
\midrule

Open Microwave
& Target opening extent
& ``Open the microwave door to [opening extent].'' \\

Open Cabinet Drawer
& Target opening extent
& ``Open the cabinet drawer to [opening extent].'' \\

Move Pill Bottle
& Placement target location
& ``Pick up the pill bottle and place it at [target location].'' \\

Move Playing Card
& Placement target location
& ``Pick up the playing card and place it at [target location].'' \\

Move Stapler
& Placement target location
& ``Pick up the stapler and place it at [target location].'' \\

\bottomrule
\end{tabularx}

\vspace{0.7em}

\begin{tabularx}{\linewidth}{
    @{}
    >{\raggedright\arraybackslash}p{0.22\linewidth}
    >{\raggedright\arraybackslash}p{0.30\linewidth}
    >{\raggedright\arraybackslash}X
    @{}
}
\toprule
\multicolumn{3}{c}{\textbf{ManiSkill}} \\
\midrule
\textbf{Task}
& \textbf{Specification}
& \textbf{Instruction Template} \\
\midrule

Reach
& End-effector target location
& ``Move the gripper to [target location].'' \\

Push Cube
& Target displacement
& ``Push the cube by [target displacement].'' \\

Pull Cube
& Object target location
& ``Grasp the cube and pull it to [target location].'' \\

Pick Cube
& Target lifting extent
& ``Lift the cube by [lifting extent].'' \\

Turn Valve
& Target rotation extent
& ``Turn the valve by [rotation extent].'' \\

\bottomrule
\end{tabularx}

\end{table}

\subsection{Dataset Construction and Sequence Sampling}
\label{app:dataset_construction}

\paragraph{Expert demonstrations.}
Training data consist exclusively of successful expert trajectories collected
using the corresponding benchmark-specific controllers. Meta-World uses its
parameter-conditioned expert controller, while RoboTwin and ManiSkill use
their respective expert or motion-planning pipelines. Only successful
trajectories are retained.

\paragraph{Data processing.}
Each trajectory stores the point-cloud observation, proprioceptive state,
action, reward, episode boundary, and task-specification metadata. Point
clouds are converted to 512 XYZ points per frame, and normalization statistics
are computed exclusively from the training demonstrations.

\paragraph{Sequence sampling.}
The same temporal sampling protocol is used across all three benchmarks.
We condition on two observation steps and use a fixed action prediction
horizon during policy training. Episode boundaries are handled using standard
boundary padding. The same sampling and preprocessing protocol is used for all
compared methods.

\subsection{Evaluation Protocol}
\label{app:evaluation_protocol}

All methods are evaluated on exactly the same held-out specifications,
initial configurations, and closed-loop rollout protocol. At each replanning
step, the policy conditions on the latest two observations, predicts an action
sequence over the configured horizon, executes the next eight actions, and
then replans. The primary evaluation metric is success rate.

Because our setting evaluates fine-grained specification following rather than
coarse task completion, we adopt stricter success thresholds than the native
benchmark criteria. For example, in Meta-World, Reach requires the final
end-effector position to be within $2.5$\,cm of the specified target, while
Push requires the planar displacement error to be below $1$\,cm. Articulated
tasks are similarly evaluated using tight errors on the specified opening
extent. All task-specific thresholds are fixed before evaluation and applied
identically to all compared methods.

Unless otherwise specified, results are aggregated over three random seeds.
Additional quantities such as final task error, return, and episode length are
retained as diagnostic metrics where supported by the corresponding
environment. Representation-level metrics, including Spearman correlation,
kNN@3, and action-probe MAE, are reported separately in
Section~\ref{app:representation_analysis}.

\subsection{Baseline Implementations and Fair Comparison}
\label{app:baselines}
\label{app:baseline-implementation}

We compare against four baselines that span three levels of language use: i) DP3, a 3D diffusion policy without any language conditioning; ii) language-conditioned 3D policies that inject language directly into the policy, \emph{3DDA} and \emph{3D-LOTUS}; iii) and T2DA, which grounds language through global behavior--instruction alignment.
All baselines are trained and evaluated on the same demonstrations, point-cloud observations, proprioceptive inputs, task instructions, and train--test specification splits, with identical held-out specifications, initial configurations, and success criteria as T3DP.
We keep each baseline's model architecture, training objective, and inference procedure unchanged, and adapt only the input and action interfaces (observation format, end-effector pose targets, and waypoint-to-action conversion) needed to place the method in our protocol.
Training budgets, training epochs, and batch sizes are aligned across methods.

\begin{list}{\textbullet}{%
  \settowidth{\labelwidth}{\textbullet}%
  \setlength{\labelsep}{0.5em}%
  \setlength{\leftmargin}{\labelwidth}%
  \addtolength{\leftmargin}{\labelsep}%
}

\item \textbf{DP3~\citep{ze2024dp3}.}
DP3 is a point-cloud diffusion policy: a lightweight MLP encoder embeds the point cloud and the proprioceptive state, and a conditional U-Net denoises a future action sequence.
It receives no language input, so specification-dependent targets are unobservable to the policy.
We use the official DP3 code directly, retaining its encoder and diffusion configuration, and train it per task on the same demonstrations as every other method, with the same prediction horizon, observation history, executed action segment, and optimization budget as our policy.

\item \textbf{3DDA~\citep{ke2024diffuseractor}.}
3D Diffuser Actor builds on a 3D feature-field transformer that lifts visual features into a 3D scene representation and applies relative-position attention over sampled action queries, and replaces single-step keypose selection with a diffusion process over end-effector pose trajectories.
Its vision--language--trajectory attention conditions the denoising network on task instructions, but supervision is still a global instruction--trajectory pair.
We adapt the official implementation only at the interface level: point clouds are mapped to the image grid expected by its 2D backbone, proprioception is converted into end-effector pose targets, the gripper command provides the openness channel, and the official task instruction is pre-encoded offline by its CLIP text tower.
Its DDPM objective, position/rotation losses, and attention structure are unchanged.
Following its native inference convention, the predicted pose segment is executed before replanning, and poses are converted into each benchmark's action space with gains calibrated from expert data.

\item \textbf{3D-LOTUS~\citep{garcia2025gembench}.}
3D-LOTUS is a point-cloud transformer policy (PTV3) that injects language through cross-attention between point features and word embeddings, and predicts actions by classification with per-axis position heatmaps, discretized rotation bins, and a binary gripper state.
We use 3D-LOTUS rather than 3D-LOTUS++, since the LLM task-planning and VLM grounding stages of the latter are not applicable to our setting, where each instruction already denotes a single grounded behavioral specification.
As with 3DDA, we adapt only the interfaces: XYZ-only point clouds without fabricated RGB channels, end-effector pose targets in a unified position--quaternion--openness format, and pose-to-increment conversion calibrated from expert data at evaluation time.
The PTV3 backbone, the cross-attention language injection, and the classification heads are unchanged.

\item \textbf{T2DA~\citep{zhang2025t2da}.}
T2DA learns a decision embedding of trajectories and aligns it with natural language through a global, pooled bidirectional contrastive objective with a learnable temperature, then conditions the downstream policy on the aligned text embedding.
We reproduce its global alignment stage and combine it with the same DP3-based diffusion policy used by T3DP.
Consequently, T2DA and T3DP share the behavior encoder and its frozen checkpoint, the text-encoder initialization, the alignment data, the downstream policy architecture, and the optimization budget, differing only in the alignment objective: T2DA optimizes the pooled global contrastive loss, whereas T3DP adds the proposed fine-grained local objective.
This comparison isolates the effect of fine-grained language-behavior alignment from changes in policy capacity, training data, or optimization.

\end{list}

\section{Implementation Details}
\label{app:implementation}

\subsection{Behavior Encoder}
\label{app:behavior_encoder}
\paragraph{Stage I: behavior representation learning.}
The behavior encoder processes a trajectory window containing point-cloud and
proprioceptive states, actions, and rewards. Point clouds are encoded by a
PointNet encoder, while the remaining modalities are projected by
modality-specific multilayer perceptrons. The state, action, and reward tokens
at each timestep share a temporal positional encoding and are processed by a
bidirectional Transformer. Mean pooling over the Transformer outputs produces
the global behavior representation. The Transformer has two layers, four
attention heads, a hidden dimension of 256, an output dimension of 256, and a
dropout rate of 0.1.

A lightweight reward decoder is trained jointly with the behavior encoder. It
contains separate projections for the action and observation features followed
by a multilayer perceptron with two hidden layers of dimension 256. The decoder
predicts the reward of a state--action pair drawn from the same task
specification but outside the encoded trajectory window. This objective
encourages the behavior representation to capture specification-dependent
behavior rather than only surface motion.

The reward signal is used only when learning the behavior representation.
Neither reward nor a demonstration is required by the deployed policy.

\subsection{Text Encoder and Language-Behavior Alignment}
\label{app:text_alignment}
\paragraph{Stage II: fine-grained Language-Behavior alignment.}
The language encoder is initialized from the CLIP ViT-B/32 text encoder. It
produces token-level representations and a pooled instruction representation.
The frozen behavior encoder provides a global behavior representation for
global alignment and local behavior tokens for fine-grained alignment. Global
alignment applies a bidirectional contrastive objective to the global behavior
and instruction representations. Local alignment applies bidirectional late
interaction between the local behavior tokens and instruction tokens. The
final alignment objective combines the global and local losses.

During alignment, the behavior encoder remains frozen and the language encoder
is adapted using LoRA on its query, key, and value projections. The LoRA rank
is 16, its scaling parameter is 32, and its dropout rate is 0.1. Instructions
are tokenized to a maximum length of 77. The local alignment dimension is 256.
The initial global contrastive temperature is 0.02 and is learnable. The local
matching temperature is 0.05, the local contrastive temperature is 0.07, and
the local alignment loss weight is 0.25. The two alignment directions are
weighted equally.

We use the terms \emph{trajectory} for a complete demonstration, \emph{behavior
window} for the fixed-length encoder input, and \emph{local alignment unit} for
the state--action hidden states retained for token-level matching.  The pooled
behavior and language representations are the corresponding global embeddings.

\subsection{Language-Conditioned 3D Policy}
\label{app:policy_implementation}
\paragraph{Stage III: language-guided 3D policy learning.}
The aligned language encoder is frozen during policy learning. The downstream
point-cloud-based 3D diffusion policy encodes the current point-cloud
observation and proprioceptive state and uses the pooled aligned language
representation as the policy condition. The denoising network predicts the diffusion noise of a noisy action sequence, conditioned on the observation history, diffusion timestep, and aligned
language representation. One policy is trained across
all training specifications within each task family. Specification-dependent
targets are not rendered in the point-cloud observation and must therefore be
communicated through language.

\subsection{Computational Resources}
\label{app:training}

Table~\ref{tab:training-hyperparameters} lists the principal training
hyperparameters used for Meta-World, RoboTwin, and ManiSkill.  Batch sizes are
reported per GPU.

\begin{table}[t]
\centering
\caption{Principal training hyperparameters across the three benchmarks.}
\label{tab:training-hyperparameters}
\small
\setlength{\tabcolsep}{5pt}
\renewcommand{\arraystretch}{1.12}
\begin{tabular}{@{}clccc@{}}
\toprule
\textbf{Stage}
& \textbf{Configuration}
& \textbf{Meta-World}
& \textbf{RoboTwin}
& \textbf{ManiSkill} \\
\midrule

\multirow{2}{*}{\textbf{I}}
& Training epochs & 500 & 500 & 500 \\
& Batch size      & 128 & 128 & 128 \\
\midrule

\multirow{2}{*}{\textbf{II}}
& Training epochs & 375 & 375 & 375 \\
& Batch size      & 128 & 128 & 128 \\
\midrule

\multirow{5}{*}{\textbf{III}}
& Training epochs       & 300 & 600 & 600 \\
& Batch size            & 512 & 128 & 512 \\
& Prediction horizon    & 16  & 16  & 16 \\
& Observation steps     & 2   & 2   & 2 \\
& Executed action steps & 8   & 8   & 8 \\
\midrule

\multirow{5}{*}{\textbf{Policy}}
& Training timesteps  & 100 & 100 & 100 \\
& Inference steps     & 10  & 10  & 10 \\
& Checkpoint interval & 25  & 50  & 50 \\
& Optimizer           & AdamW & AdamW & AdamW \\
& Learning rate       & $10^{-4}$ & $10^{-4}$ & $10^{-4}$ \\
\bottomrule
\end{tabular}
\end{table}

All stages use AdamW with weight decay $10^{-6}$. The policy uses a DDIM noise
scheduler. Point-cloud observations contain 512 XYZ points and do not include
RGB features.

We train our models on one NVIDIA GeForce RTX 3090 GPU with dual AMD EPYC
9654 CPUs and 512 GB RAM. Behavior-encoder pre-training, fine-grained
language-behavior alignment, and language-guided policy training each cost
about 1 hour, with slight variations depending on the complexity of the
environment and data-loading overhead. Consequently, the complete three-stage
training pipeline takes approximately 3 hours.

\clearpage

\section{Training and Inference Algorithms}
\label{app:alg}
\label{app:algorithms}

T3DP is trained in three stages.  The following pseudocodes make explicit
which components are optimized in each stage and clarify that deployment uses
only the aligned text encoder and the policy.

\subsection{Behavior Representation Learning}
\label{app:behavior_algorithm}

Let a length-$L$ trajectory window be
$\tau_w=\{(s_t,a_t,r_t)\}_{t=w}^{w+L-1}$, where $s_t$ contains the point-cloud
observation and proprioceptive state.  The behavior encoder tokenizes the
state, action, and reward at every timestep, processes the resulting $3L$
tokens with a bidirectional Transformer, and mean-pools its outputs.

\subsection{Global and Local Language-Behavior Alignment}
\label{app:alignment_algorithm}

For an instruction $l$, the text encoder produces token representations
$E(l)=\{e_j\}_{j=1}^{M}$ and the pooled representation
$c_l=M^{-1}\sum_{j=1}^{M}e_j$.  For its paired trajectory $\tau$, let
$H_\varphi(\tau)$ denote the hidden states of the frozen behavior Transformer.
The global behavior representation is
$c_\tau=\operatorname{MeanPool}(H_\varphi(\tau))$, while
$Z(\tau)=\{z_i\}_{i=1}^{N}$ retains its state and action hidden states for
local matching.  With $s_{ij}=\cos(z_i,e_j)$, define
\begin{align}
S_{\tau\rightarrow l}(Z,E)
&=\frac{1}{N}\sum_{i=1}^{N}
\operatorname{LME}_{\beta}\!\left(\{s_{ij}\}_{j=1}^{M}\right),\\
S_{l\rightarrow\tau}(Z,E)
&=\frac{1}{M}\sum_{j=1}^{M}
\operatorname{LME}_{\beta}\!\left(\{s_{ij}\}_{i=1}^{N}\right),
\label{eq:app-local-scores}
\end{align}
where
$\operatorname{LME}_{\beta}(\{x_k\}_{k=1}^{K})
=\beta^{-1}\log\!\left(K^{-1}\sum_{k=1}^{K}\exp(\beta x_k)\right)$.
Both the global similarities and the two local scores are optimized with
bidirectional contrastive objectives over paired trajectories and
instructions.

\subsection{Language-Conditioned Policy Learning}
\label{app:policy_algorithm}

The aligned text encoder is frozen during policy learning.  Following the
method definition in the main paper, the downstream policy consumes only the
pooled aligned representation $c_l$; the local objective shapes this pooled
condition but does not pass token-level features directly to the policy.

\subsection{Inference without Demonstrations}
\label{app:inference}

At deployment, the behavior encoder, rewards, and demonstrations are not used,
and the aligned text encoder is not updated.  The policy receives only the
current point cloud, robot state, and language instruction; it samples a
horizon-length action sequence, executes the configured action segment, and
then replans from the next observation.  Consequently, a held-out
specification does not require a corresponding training demonstration at test
time.

\clearpage

\begin{algorithm}[H]
\caption{Point-cloud behavior representation learning}
\label{alg:behavior-learning}

\KwIn{
Expert demonstration sets $\{\mathcal D_\sigma\}_{\sigma}$;
behavior encoder $e_\varphi$;
reward decoder $d_\psi$
}
\KwOut{Trained behavior encoder $e_\varphi$}

\ForEach{training epoch}{
    \ForEach{specification $\sigma$, trajectory $\tau\in\mathcal D_\sigma$,
    and sampled window $\tau_w\subset\tau$}{
        Encode point cloud and proprioception as state tokens, and embed the
        corresponding actions and rewards\;

        Add temporal encodings and form the interleaved
        state--action--reward token sequence $X_w$\;

        $c_\tau \leftarrow e_\varphi(\tau_w)$ by processing $X_w$
        with the bidirectional Transformer and mean pooling\;

        Sample $(s,a,r)\in\tau\setminus\tau_w$\;

        $\hat r \leftarrow d_\psi(c_\tau,s,a)$\;

        Update $(\varphi,\psi)$ using
        $\widehat{\mathcal L}_{\mathrm{behavior}}
        =(\hat r-r)^2$\;
    }

    Evaluate reward prediction on the validation split\;
}

Retain the selected behavior-encoder checkpoint\;

\end{algorithm}

\begin{algorithm}[H]
\caption{Global--local Language-Behavior alignment}
\label{alg:language-behavior-alignment}

\KwIn{
Frozen behavior encoder $e_\varphi$;
paired trajectory--instruction minibatches
$\{(\tau_b,l_b)\}_{b=1}^{B}$;
text encoder $g_\phi$ with LoRA;
text-side projections;
local-loss weight $\lambda$
}
\KwOut{
Aligned text encoder $g_\phi$ and text-side projections
}

\ForEach{paired minibatch $\{(\tau_b,l_b)\}_{b=1}^{B}$}{

    $\{(c_{\tau_b},Z_b)\}_{b=1}^{B}
    \leftarrow
    e_\varphi(\{\tau_b\}_{b=1}^{B})$\;

    Retain global behavior embeddings $c_{\tau_b}$ and local
    state--action representations $Z_b$\;

    $\{(c_{l_b},E_b)\}_{b=1}^{B}
    \leftarrow
    g_\phi(\{l_b\}_{b=1}^{B})$\;

    Retain pooled instruction embeddings $c_{l_b}$ and token
    representations $E_b$\;

    Compute $\mathcal L_{\mathrm{global}}$ from global
    language-behavior similarities\;

    Compute $\mathcal L_{\mathrm{local}}$ from
    $S_{\tau\rightarrow l}(Z_b,E_q)$ and
    $S_{l\rightarrow\tau}(Z_q,E_b)$ using
    Eq.~\eqref{eq:app-local-scores}\;

    $\mathcal L_{\mathrm{align}}
    \leftarrow
    \mathcal L_{\mathrm{global}}
    +
    \lambda\mathcal L_{\mathrm{local}}$\;

    Update the LoRA parameters of $g_\phi$ and the text-side projections
    using $\mathcal L_{\mathrm{align}}$\;
}

\end{algorithm}

\begin{algorithm}[H]
\caption{Language-conditioned diffusion policy learning and inference}
\label{alg:language-conditioned-policy}

\KwIn{
Policy training tuples
$\mathcal D_\pi=\{(o_t,p_t,l,\mathbf A^0)\}$;
frozen aligned text encoder $g_\phi$;
observation encoder $F_\eta$;
denoiser $\epsilon_\omega$;
DDIM scheduler
}
\KwOut{
Trained policy parameters $(\eta,\omega)$
}

\BlankLine
\textbf{Training}

\ForEach{minibatch
$(o_t,p_t,l,\mathbf A^0)\sim\mathcal D_\pi$}{

    $c_l \leftarrow \operatorname{MeanPool}(g_\phi(l))$\;

    $h_t \leftarrow F_\eta(o_t,p_t)$\;

    Sample diffusion step $k$ and
    $\epsilon\sim\mathcal N(0,I)$\;

    Construct $\mathbf A^k$ using the forward diffusion scheduler\;

    $\hat\epsilon
    \leftarrow
    \epsilon_\omega(\mathbf A^k,k,h_t,c_l)$\;

    Update $(\eta,\omega)$ using
    $\mathcal L_{\mathrm{policy}}
    =
    \lVert\epsilon-\hat\epsilon\rVert_2^2$\;
}

\BlankLine
\textbf{Receding-horizon inference}

Given the current $(o_t,p_t,l)$, compute $c_l$ and $h_t$ as above\;

Initialize $\mathbf A^K\sim\mathcal N(0,I)$\;

\For{$k=K,K-1,\ldots,1$}{

    $\hat\epsilon
    \leftarrow
    \epsilon_\omega(\mathbf A^k,k,h_t,c_l)$\;

    Obtain $\mathbf A^{k-1}$ with one DDIM reverse step\;
}

Unnormalize $\mathbf A^0$\;

Execute the configured action segment and replan from the next observation\;

\end{algorithm}

\clearpage

\clearpage

\section{Additional Experimental Results}
\raggedbottom
\label{app:additional_results}

\subsection{Additional RoboTwin Results}
\label{app:robotwin_results}

\paragraph{Experimental setting.}
We further evaluate unseen-specification generalization on five RoboTwin task
families: \emph{Open Microwave}, \emph{Open Cabinet Drawer},
\emph{Move Pill Bottle}, \emph{Move Playing Card}, and
\emph{Move Stapler}. These tasks cover both articulated-object manipulation
and spatial object placement. We compare DP3 without language conditioning,
direct conditioning with pretrained CLIP representations (Raw CLIP), T2DA
with global language-behavior alignment, and T3DP with the proposed
fine-grained alignment. All methods use the same demonstrations, observation
inputs, language instructions, train--test specification splits, and
evaluation protocol. We evaluate exclusively on held-out specifications
following the protocol described in Appendix~\ref{app:task_details}.

\begin{table}[H]
\centering
\caption{
Unseen-specification generalization on RoboTwin.
We report success rates (\%) and corresponding standard deviations over three
independently trained seeds. ``Avg.'' denotes the mean across the five task
families. Best and second-best results are shown in \textbf{bold} and \underline{underlined},
respectively.
}
\label{tab:robotwin_results}
\small
\setlength{\tabcolsep}{2.8pt}
\begin{tabular}{lcccccc}
\toprule
Method
& \makecell{Open\\Microwave}
& \makecell{Open Cabinet\\Drawer}
& \makecell{Move Pill\\Bottle}
& \makecell{Move Playing\\Card}
& \makecell{Move\\Stapler}
& Avg. \\
\midrule
DP3
& 15.7\ensuremath{\pm}8.6
& 5.0\ensuremath{\pm}1.3
& 3.7\ensuremath{\pm}1.3
& 9.0\ensuremath{\pm}2.2
& 2.7\ensuremath{\pm}2.4
& 7.2 \\

Raw CLIP
& 37.7\ensuremath{\pm}9.5
& 12.3\ensuremath{\pm}10.9
& 19.3\ensuremath{\pm}3.8
& 28.7\ensuremath{\pm}3.4
& 22.0\ensuremath{\pm}2.8
& 24.0 \\

T2DA
& \underline{46.0\ensuremath{\pm}4.5}
& \underline{16.7\ensuremath{\pm}16.3}
& \underline{34.3\ensuremath{\pm}1.3}
& \underline{32.7\ensuremath{\pm}2.5}
& \underline{24.3\ensuremath{\pm}1.7}
& \underline{30.8} \\

\textbf{T3DP (Ours)}
& \textbf{52.0\ensuremath{\pm}13.5}
& \textbf{37.3\ensuremath{\pm}0.6}
& \textbf{39.3\ensuremath{\pm}0.5}
& \textbf{52.3\ensuremath{\pm}1.5}
& \textbf{28.0\ensuremath{\pm}1.6}
& \textbf{41.8} \\
\bottomrule
\end{tabular}
\end{table}

\paragraph{Result analysis.}
T3DP achieves an average success rate of 41.8\%, compared with 30.8\% for
T2DA, 24.0\% for Raw CLIP, and 7.2\% for DP3. Fine-grained alignment therefore
improves over the global-alignment baseline by 11.0 percentage points.
Importantly, T3DP improves over T2DA on all five RoboTwin task families,
including both articulated-object manipulation and object-placement tasks.
The gains are particularly pronounced on \emph{Open Cabinet Drawer} and
\emph{Move Playing Card}, where T3DP improves success from 16.7\% to 37.3\%
and from 32.7\% to 52.3\%, respectively.

The consistent improvement over Raw CLIP and T2DA provides additional evidence
that behavior grounding is beneficial beyond direct language conditioning,
while fine-grained language-behavior correspondence further improves
generalization beyond global alignment. Together with the Meta-World and
ManiSkill results, these results show that the advantage of T3DP transfers
across different manipulation environments, embodiments, and action spaces.

\subsection{Alignment Representation Analysis}

\label{app:representation_analysis}
\label{app:representation}
Beyond downstream success, we examine what the learned alignment preserves
and how this structure relates to control. We focus on two complementary views:
the geometry of the alignment space and the action relevance of the learned
language condition.

\paragraph{Representation geometry.}
We evaluate whether the learned language representations preserve the
underlying specification structure on Reach, Push, and Door Open, covering
spatial targets, relative displacements, and articulated states.
We report Spearman correlation between pairwise specification and
representation distances, together with kNN@3 neighborhood preservation.

\begin{table}[t]
\centering
\caption{
Representation geometry on held-out Meta-World specifications.
}
\label{tab:alignment_metrics}
\small
\setlength{\tabcolsep}{5.5pt}
\begin{tabular}{llccc}
\toprule
Task & Metric & Generic & Global & Fine-grained \\
\midrule
\multirow{2}{*}{Reach}
& Spearman $\uparrow$ & 0.504 & 0.567 & \textbf{0.612} \\
& kNN@3 $\uparrow$    & 0.417 & 0.450 & \textbf{0.533} \\
\midrule
\multirow{2}{*}{Push}
& Spearman $\uparrow$ & 0.255 & 0.524 & \textbf{0.708} \\
& kNN@3 $\uparrow$    & 0.283 & 0.317 & \textbf{0.567} \\
\midrule
\multirow{2}{*}{Door Open}
& Spearman $\uparrow$ & 0.360 & 0.682 & \textbf{0.841} \\
& kNN@3 $\uparrow$    & 0.400 & 0.433 & \textbf{0.622} \\
\bottomrule
\end{tabular}
\end{table}

Global alignment improves specification structure over generic language
features, while fine-grained alignment further improves both global ordering
and local neighborhood preservation across all three task types.
This supports our hypothesis that finer alignment better preserves
specification-dependent behavioral variation.

\paragraph{Action relevance of the learned language condition.}
Representation geometry alone does not establish whether the learned structure
is informative for control.
We therefore construct a \emph{same-state action probe}, where the robot and
scene state are fixed while the language condition varies across held-out
specifications.
This removes state-dependent shortcuts, such that differences in the target
actions must be explained by specification-dependent information encoded in
the language representation.
Probe training and evaluation use disjoint source trajectories, and results
are averaged over four trajectory splits.

\begin{figure}[t]
    \centering
    \includegraphics[width=0.94\linewidth]{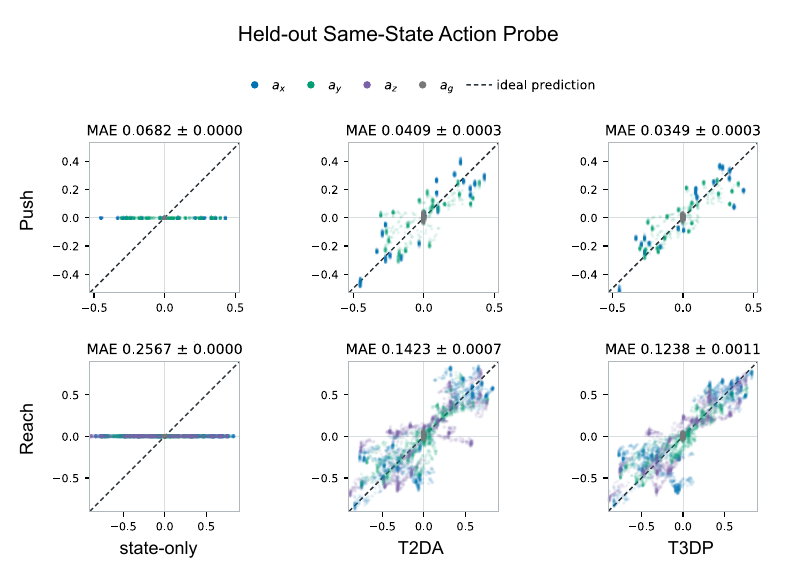}
    \caption{
    Full same-state action probe results on held-out Push and Reach
    specifications.
    Each point compares the predicted and target action under a fixed state
    while varying the language specification; the dashed diagonal denotes
    ideal prediction.
    State-only predictions collapse toward the mean action, whereas
    behavior--language alignment recovers specification-dependent action
    variation, with T3DP more closely following the target actions than
    global alignment (T2DA).
    }
    \label{fig:action_probe_full}
\end{figure}

As shown in Figure~\ref{fig:action_probe_full}, conditioning on an aligned
language representation substantially reduces action error relative to the
state-only probe, indicating that behavior--language alignment captures
information about how different specifications require different actions.
Fine-grained alignment further reduces MAE over global alignment by
$14.7\%$ on Push and $13.0\%$ on Reach.
The consistent trend across relative-displacement and spatial-target
specifications provides complementary evidence that T3DP better preserves
specification-dependent action information in the condition consumed by the
downstream policy.
Given the limited number of held-out specifications, we treat this probe as
diagnostic evidence rather than a standalone statistical claim.

\section{Real-Robot Experimental Details}
\label{app:real_robot}

\subsection{Robot and Sensor Setup}

\begin{wrapfigure}{r}{0.30\textwidth}
\vspace{-1.2em}
\centering
\includegraphics[width=\linewidth]{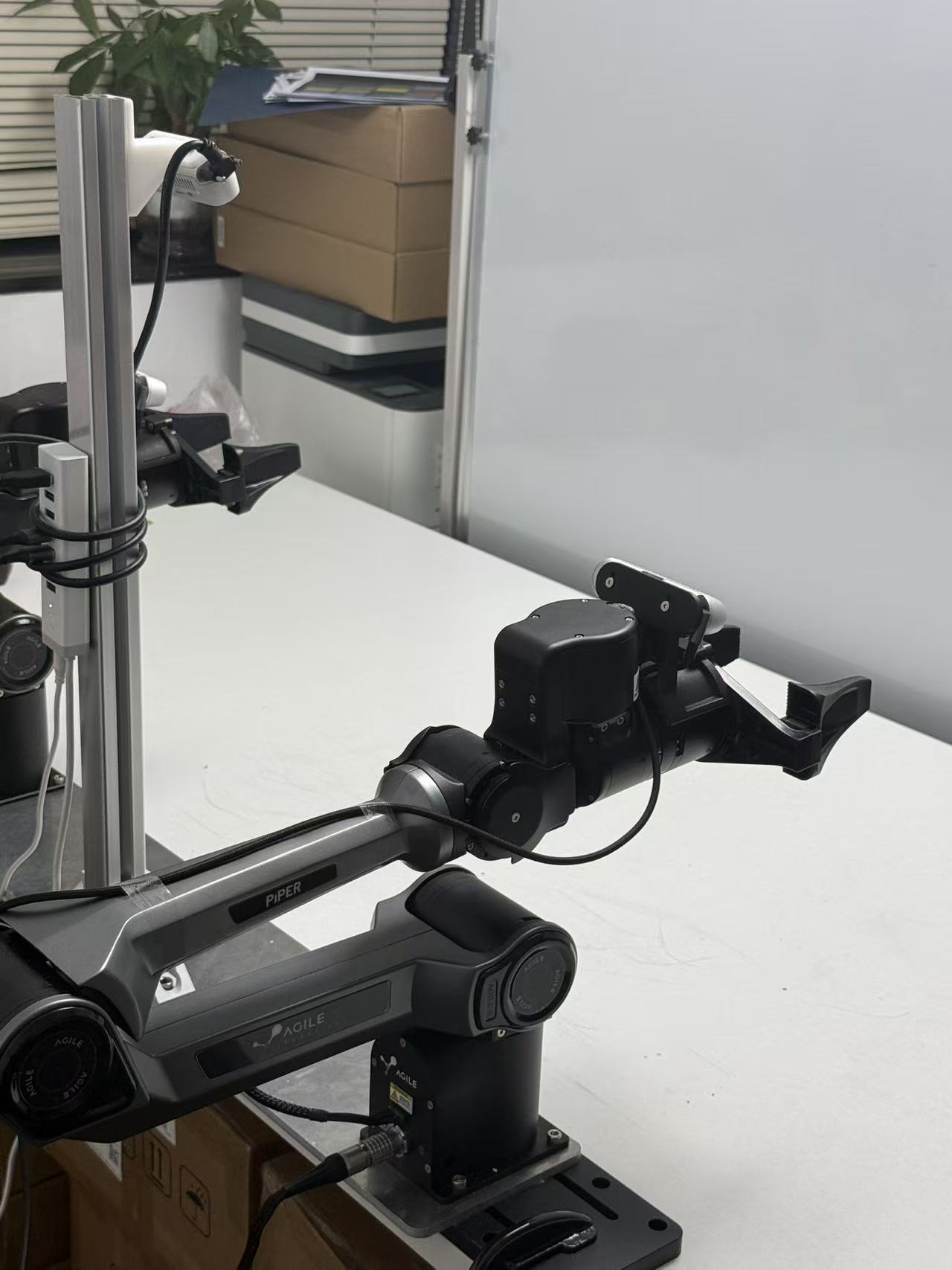}
\captionsetup{
    font=footnotesize,
    justification=raggedright,
    singlelinecheck=false,
    skip=1pt
}
\caption{
Real-robot setup with a PiPer X manipulator and D435i depth camera.
}
\label{fig:real_robot_setup}
\vspace{-4em}
\end{wrapfigure}

We conduct real-robot experiments using a PiPer X manipulation platform.
Although the platform supports dual-arm operation, all experiments reported
here use a single PiPer X manipulator.
An Intel RealSense D435i depth camera is mounted above the workspace and
remains fixed throughout data collection and evaluation.
Depth observations are transformed into the robot coordinate frame using a
fixed extrinsic calibration.

The policy receives geometric point-cloud observations and robot
proprioceptive states; RGB information is not used.
We retain the same observation modality and end-effector action
parameterization as in simulation.
Specification-dependent target information is not explicitly encoded in the
visual observation and is instead communicated to the policy through language.

\subsection{Point-Cloud Construction}
\label{app:real_point_cloud}

Depth images are back-projected to 3D and transformed into the robot
coordinate frame. Points outside the task workspace are removed, after which
the remaining cloud is subsampled to 512 XYZ points using the same input
format as in simulation. No RGB features are provided to the policy.
The same normalization convention is used for simulation and real-robot
experiments.

\subsection{Tasks and Target Specifications}

We evaluate two placement tasks, \emph{Move Pill Bottle} and
\emph{Move Stapler}. In both tasks, the specification is defined by a target
placement location in the tabletop plane and communicated to the policy
through a fixed language template.

\begin{figure}[t]
    \centering
    \begin{minipage}[t]{0.49\linewidth}
        \centering
        \includegraphics[width=\linewidth]{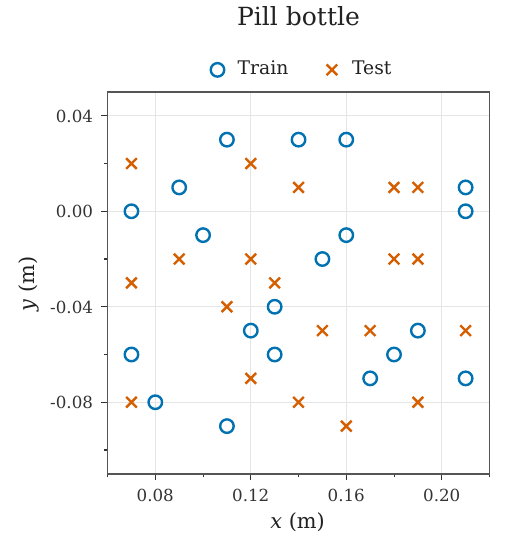}
    \end{minipage}
    \hfill
    \begin{minipage}[t]{0.49\linewidth}
        \centering
        \includegraphics[width=\linewidth]{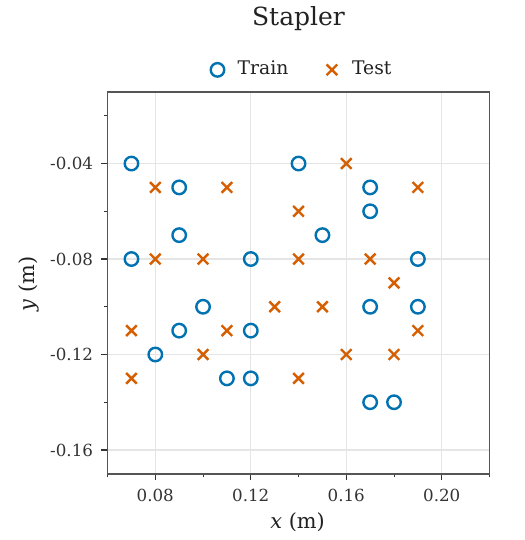}
    \end{minipage}
    \caption{
    Spatial distribution of the target specifications used in the real-robot
    experiments. Left: \emph{Move Pill Bottle}. Right: \emph{Move Stapler}.
    Blue markers denote training specifications, orange markers denote
    held-out specifications. Each task uses 20 training and 20 held-out target
    specifications.
    }
    \label{fig:real_goal_distribution}
\end{figure}

For each task, we use 20 training target specifications and 20 disjoint
held-out target specifications. Figure~\ref{fig:real_goal_distribution}
visualizes their spatial distributions. The held-out targets cover the same
feasible placement region as the training targets while remaining unseen
during training, allowing us to evaluate generalization to unseen target
specifications without introducing an additional shift in workspace coverage.

\subsection{Real-Robot Demonstration Collection}
\label{app:real_data_collection}

Real-robot demonstrations are collected through a simulation-assisted
pipeline. For each task specification, we first instantiate the corresponding
task in a modified RoboTwin environment and use the simulator expert to
generate a successful reference trajectory. The resulting trajectory is then
replayed on the real PiPer X robot, during which real depth observations,
robot proprioception, and executed actions are recorded to construct the
training demonstration.

The simulator is used only to generate the expert motion sequence; policy
training uses the observations collected during real-robot execution.
Only successful real-world replays are retained in the training dataset.

\subsection{Execution and Success Criteria}
\label{app:real_execution}

We compare DP3 without language conditioning, T2DA with global
language-behavior alignment, and T3DP under the same task and evaluation
settings. All methods are evaluated on the same 20 held-out target
specifications for each task. Each held-out specification is executed once,
resulting in 20 real-robot trials per task and method.

A trial is considered successful when the final object position is within
$3\,\mathrm{cm}$ of the specified target in the tabletop plane.
Target locations are used only for evaluation and are not explicitly encoded
in the policy observation. Quantitative success rates are reported in the
main paper.

\subsection{Qualitative Results and Failure Cases}
\label{app:real_qualitative}

DP3 produces final object positions farthest from the targets.  T2DA typically
brings objects closer, whereas T3DP yields the strongest placement precision
and success rate across the two task families.  The qualitative examples in
Figure~\ref{fig:real_world} illustrate the remaining failures as grasping or
placement errors, including cases that approach but do not meet the 3\,cm
success threshold.

\section{Limitations}
\label{app:limitations}

Our behavior encoder is trained with a reward-predictive objective and therefore assumes access to expert trajectories annotated with rewards, which are often unavailable or costly to define in real-world datasets, and many existing robot datasets provide only state--action sequences without a reliable reward signal.
A plausible remedy is to learn specification-dependent behavior representations directly from state--action sequences through self-supervised objectives, which we leave to future work.
Our evaluation focuses on held-out specifications within the task families seen during training, and our instructions are instantiated from fixed task-specific templates.
We additionally study mixed-task training in
Section~\ref{sec:scalability}, where a single policy is trained jointly
across five Meta-World task families, but that study is limited in scale
and confined to one benchmark.
Scaling joint training to a larger and more heterogeneous collection of tasks, and extending T3DP to unseen task families, longer-horizon behaviors, and more diverse embodiments, therefore remains an important direction for future work.

\end{document}

%% file: math_commands.tex
\usepackage{amsmath,amsfonts,bm}

\def\eqref#1{\textup{(\ref{#1})}}

\def\1{\bm{1}}

\DeclareMathAlphabet{\mathsfit}{\encodingdefault}{\sfdefault}{m}{sl}
\SetMathAlphabet{\mathsfit}{bold}{\encodingdefault}{\sfdefault}{bx}{n}

